\documentclass{article}
\usepackage{spconf,amsmath,graphicx}
\usepackage{booktabs,multirow, multicol}

\newcommand{\diu}{DIU{} }

\newcommand{\fulldiu}{Domain Invariant Units{} }
\usepackage{xcolor}
\usepackage{colortbl}
\usepackage{hyperref}

\definecolor{Gray}{rgb}{0.9, 0.9, 0.9}%https://latexcolor.com/

\newcommand{\hfr}{\textit{HFR}{} }

\title{Heterogeneous Face Recognition Using Domain Invariant Units}

\name{Anjith George$^1$ and S\'{e}bastien Marcel$^{1,2}$
\thanks{This research is based upon work supported in part by the Office of the Director of National Intelligence (ODNI), Intelligence Advanced Research Projects Activity (IARPA), via [2022-21102100007]. The views and conclusions contained herein are those of the authors and should not be interpreted as necessarily representing the official policies, either expressed or implied, of ODNI, IARPA, or the U.S. Government. The U.S. Government is authorized to reproduce and distribute reprints for governmental purposes notwithstanding
 any copyright annotation therein.}
}
\address{$^1$Idiap Research Institute, Martigny, Switzerland\\
$^2$University of Lausanne, Switzerland\\
{\tt\small \{anjith.george,sebastien.marcel\}@idiap.ch}
}

\begin{document}
%\ninept
%
\maketitle
\begin{abstract}

Heterogeneous Face Recognition (HFR) aims to expand the applicability of Face Recognition (FR) systems to challenging scenarios, enabling the matching of face images across different domains, such as matching thermal images to visible spectra. However, the development of HFR systems is challenging because of the significant domain gap between modalities and the lack of availability of large-scale paired multi-channel data. In this work, we leverage a pretrained face recognition model as a teacher network to learn domain-invariant network layers called Domain-Invariant Units (DIU) to reduce the domain gap. The proposed DIU can be trained effectively even with a limited amount of paired training data, in a contrastive distillation framework. This proposed approach has the potential to enhance pretrained models, making them more adaptable to a wider range of variations in data. We extensively evaluate our approach on multiple challenging benchmarks, demonstrating superior performance compared to state-of-the-art methods.
\end{abstract}
\section{Introduction}
\label{sec:intro}

Face recognition (FR) has gained popularity as an access control tool due to its effectiveness and user-friendly nature. State-of-the-art FR methods often demonstrate outstanding performance in real-world conditions \cite{learned2016labeled}. Nevertheless, these methods fail when applied in challenging situations such as low light conditions or over long distances. Typically, FR systems operate in a homogeneous domain, enrolling and matching individuals using the same modality—usually employing facial images captured through an RGB camera. This approach proves inadequate when facing scenarios where enrollment and the probe images exhibit heterogeneity. 

\begin{figure}[t!]
  \centering
  \includegraphics[width=0.85\linewidth, trim=0 0 120 0]{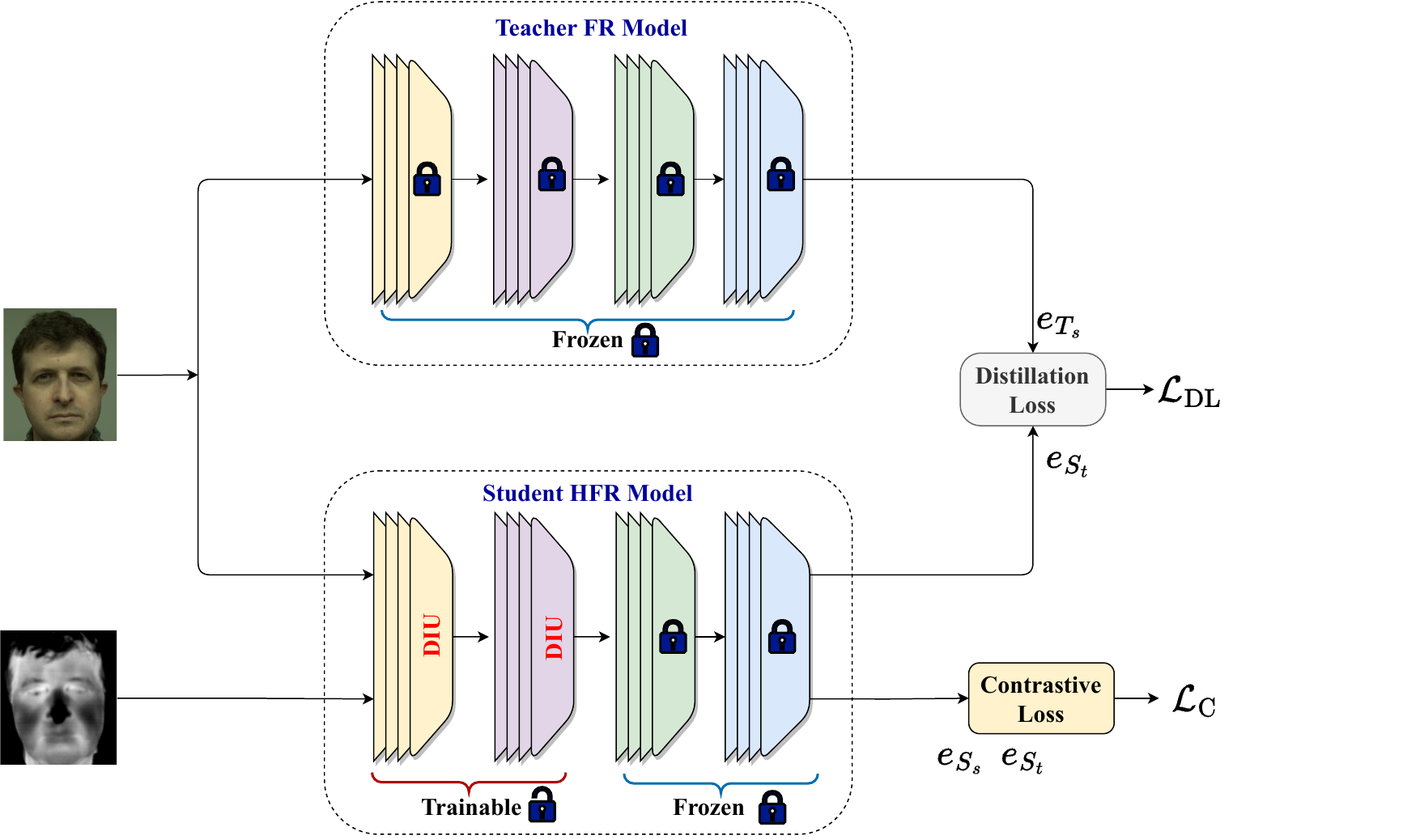}
  \caption{The proposed Domain Invariant Unit (DIU) framework. The lower layers of the student model are trained in a contrastive framework to learn invariant features, while supervision from the distillation loss prevents overfitting.}
  \vspace{-5mm}
  \label{fig:framework}
\end{figure}

There are several instances where face matching in heterogeneous environments can prove beneficial. Consider a scenario in which the images used for enrollment originate from a controlled setting, while the target images come from a CCTV camera utilizing the Near Infrared (NIR) spectrum. NIR images can be used for both day and night regardless of illumination conditions. Similarly, the concept of matching images across different domains, such as visible and thermal images, proves beneficial in cases where the use of active illumination is not feasible. It's worth noting that although the term ``heterogeneity'' often refers to various modalities in the literature, discrepancies such as alterations in image quality (as seen in long-distance recognition) can also introduce heterogeneity. Heterogeneous Face Recognition (HFR) systems aim to facilitate cross-domain matching, allowing RGB images used for enrollment to be compared with images from any diverse modality. This obviates the necessity for separate modalities to undergo enrollment \cite{klare2012heterogeneous}. HFR is especially advantageous in scenarios where capturing good-quality visible images is challenging.

While being highly beneficial, achieving \hfr using FR models trained using large-scale RGB datasets is challenging due to the domain gap \cite{he2018wasserstein}. Moreover, the limited availability of paired datasets with the target modalities makes training invariant models from scratch challenging. Gathering large data for these new modalities can be cost-prohibitive. As a result, it's important to devise a framework that requires only a minimal set of paired data samples.

Several approaches have been proposed for heterogeneous face recognition (\hfr\!) to extract invariant features that can match cross-modal images. Feature-based methods such as Difference of Gaussian (DoG) \cite{liao2009heterogeneous}, and scale-invariant feature transform (SIFT) \cite{klare2010matching} have been proposed for \hfr to reduce the domain gap. Another set of methods called common-space projection methods aims to learn a mapping to project different face modalities into a common shared subspace to reduce the domain gap \cite{kan2015multi,he2017learning}. Recent HFR methods \cite{fu2021dvg,zhang2017generative}, primarily use GAN-based synthesis due to its high-quality image generation. While using pre-trained FR models in synthesis-based HFR reduces the data needed for a new FR model's training, the synthesis increases computational costs, limiting real-world applicability. In \cite{de2018heterogeneous}, a new HFR method called Domain-Specific Units (DSU) was introduced. The authors argue that high-level features of CNNs trained on the visible spectrum can encode images from other modalities, indicating these features are domain-independent. Prepended Domain Transformers (PDT) \cite{george2022prepended} adds a separate module to a pretrained FR network for the target modality, converting it to an HFR network. In \cite{george2023bridging}, authors treat different modalities as different styles and try to match them by modulating feature maps of the target modality using a conditional adaptive instance modulation (CAIM).
In this work, in contrast to domain specific units (DSU), we propose to learn domain invariant units (DIU). We achieve this in a teacher-student distillation learning setting, adapting the lower network layers with a minimal set of paired samples.
The main contributions of this work are listed below:
  \begin{itemize}
    \item We formulate the Heterogeneous Face Recognition (\hfr\!) problem in a teacher-student distillation framework, leveraging a pretrained FR system as the teacher.
    \vspace{-5.5mm}
    \item The proposed approach can be used to improve pretrained models to more variability given a minimal amount of paired data.
    
    \item We have evaluated the proposed approach in several standard benchmarks to show the effectiveness of our approach.   
  \end{itemize}
  
Finally, the protocols and source codes will be made available publicly \footnote{\url{https://gitlab.idiap.ch/bob/bob.paper.icassp2024_diu_hfr}}.

\section{Proposed Method}
\label{sec:approach}
%============================================================================================================================================================

Several recent works \cite{de2018heterogeneous,george2022prepended} have shown that conditionally adapting low-level features can improve HFR performance. For instance, Domain Specific Units (DSU) \cite{de2018heterogeneous} proposes to create a copy of the initial layers from a Face Recognition (FR) network and adapt them specifically for the target modality. Prepended Domain Transfomers (PDT) on the other hand add a prepended module to the FR network specifically for the target modality. While these approaches perform well, they introduce an asymmetry in the data flow. The path for the source modality is always fixed (a pretrained FR network trained on a large face dataset), meaning the HFR training process has to always learn to map to this predefined space. This paradigm implies that the HFR training regime is invariably directed towards this pre-defined space. Such a constraint potentially curtails the training framework's capability to learn optimal invariant features. While training a single model for both source and target modality is possible, this will likely result in overfitting, especially given the limited volume of data typically found in heterogeneous datasets.

We propose to address this challenge by fine-tuning components of a pretrained face recognition network in a domain-invariant manner. More specifically, we aim to make the framework more flexible so that the source and target modality can align well in the embedding space by learning invariant layers. To counteract overfitting, we adopt a dual-strategy approach: 1) Restricting the training process to a specified subset of the lower layers, and 2) Incorporating supervisory signals from a teacher network, thereby safeguarding against catastrophic forgetting.

Let $T$ denote a pretrained face recognition network with parameters given by $\Theta_{FR}$, trained on a large-scale dataset. Let heterogeneous image pairs from the source and target modalities be represented by $X_{s_{i}}$ and $X_{t_{i}}$ respectively. The label $y_i$ signifies if the pairs correspond to the same identity (1) or not (0). Given this, we define the embeddings from the model $T$ for the source and target modalities as $e_{T_{s_{i}}}$ and $e_{T_{t_{i}}}$, respectively. In our formulation, the objective is to learn a \textit{student} network $S$ parameterized by $\Theta_{HFR}$, such that the embeddings obtained for $X_{s_{i}}$ and $X_{t_{i}}$ belonging to the same identity ($y_i=1$), $e_{S_{s_{i}}}$ and $e_{S_{t_{i}}}$ align in the embedding space. 

Note that the values of $\Theta_{HFR}$ is initialized with $\Theta_{FR}$. Now, we represent the weights of $\Theta_{HFR}$ into two categories. 
\begin{equation}
\Theta_{HFR}= \{\Theta_{DIU_{i=1,2,..,K}}, \Theta_{Frozen}\}
\end{equation}
Let $\Theta_{Frozen}$ represent the layers that remain unchanged during training, and $\Theta_{DIU_{i}}$ represent the $i^{th}$ \diu block out of a total of $K$ blocks that have been trained to learn an invariant representation for both the source and target modalities.

To align the representation we use a cosine contrastive loss function, denoted by $\mathcal{L}_{C}$:
\[
\resizebox{.99\linewidth}{!}{$
\begin{aligned}
\mathcal{L}_{C}(e_{S_{s_{i}}}, e_{S_{t_{i}}}, y_i) = & (1 - y_i) \cdot \max\left(0, \frac{e_{S_{s_{i}}} \cdot e_{S_{t_{i}}}}{\| e_{S_{s_{i}}} \|_2 \| e_{S_{t_{i}}} \|_2} - m\right) \\
& +  y_i \cdot \left(1 - \frac{e_{S_{s_{i}}} \cdot e_{S_{t_{i}}}}{\| e_{S_{s_{i}}} \|_2 \| e_{S_{t_{i}}} \|_2}\right)\\
\end{aligned}
$}
\]

Where $m$ denotes the margin. 

To ensure the stability of network training and prevent overfitting to the limited \hfr data, we enforce a criterion such that the embeddings generated by the \textit{student} network for the source modality images should match those produced by a teacher network. This approach resembles a distillation framework \cite{hinton2015distilling}, wherein the student network is guided to mimic the representation learned by the teacher \cite{romero2014fitnets, tian2019contrastive}. In this context, it can be seen as a form of self-distillation; however, it is applied exclusively to images from the source modality and is focused solely on adapting the DIU layers.

Here the distillation loss has the following form $\mathcal{L}_{DL}$:

\begin{equation}
\mathcal{L}_{DL}(e_{T_{s_{i}}}, e_{S_{s_{i}}}) = \|e_{T_{s_{i}}}- e_{S_{s_{i}}}\|_2 
\end{equation}

Now the combined loss function to optimize can be written as:
\begin{equation}
  \resizebox{.9\linewidth}{!}{$
  \begin{aligned}
  \mathcal{L}_{C}(e_{S_{s_{i}}}, e_{S_{t_{i}}}, y_i) = & (1 - y_i) \cdot \max\left(0, \frac{e_{S_{s_{i}}} \cdot e_{S_{t_{i}}}}{\| e_{S_{s_{i}}} \|_2 \| e_{S_{t_{i}}} \|_2} - m\right) \\
  & +  y_i \cdot \left(1 - \frac{e_{S_{s_{i}}} \cdot e_{S_{t_{i}}}}{\| e_{S_{s_{i}}} \|_2 \| e_{S_{t_{i}}} \|_2}\right)\\
  \end{aligned}
  $}
  \end{equation}

where $\gamma$ is a hyper-parameter to determine the contribution of each component in the total loss. The value of $\gamma$ is empirically selected as 0.75, and the value of $m$ is set as zero in all the experiments. 

\textbf{Pre-trained FR backbone}: We employed the pre-trained \textit{Iresnet101} face recognition model from AdaFace \cite{kim2022adaface}. Specifically, the model we used was trained using the WebFace12M dataset \cite{zhu2021webface260m}, which contains over 12M images representing more than 600K identities. The model processes three-channel images with a resolution of $112 \times 112$ pixels. Before inputting into the Face Recognition network, the faces are aligned and cropped to align the eye centers with specific coordinates. For single-channel thermal images, the data is duplicated thrice during preprocessing.

\textbf{Implementation details}: The proposed \fulldiu (DIU) is trained using a teacher-student distillation framework, complemented by a contrastive learning component. This framework was implemented using PyTorch and integrated with the Bob library \cite{bob2017,bob2012} \footnote{\url{https://www.idiap.ch/software/bob/}}. We employed the Adam Optimizer, with a learning rate of $0.0001$, and trained the framework for 50 epochs with a batch size of 48. The margin parameter $m$ at zero and hyperparameter $\gamma$ to 0.75 in all the experiments. We first initialize the student HFR models using the weights from the teacher. Only a certain number of lower layers of the student model are set as trainable, keeping the remaining ones frozen. The source channel image embeddings, obtained from the frozen teacher network, along with the contrastive loss acting on the student HFR network force the trainable layers to learn invariant representations for both modalities. During inference, solely the student network is required for performing the HFR task.

\section{Experiments}
\label{sec:experiments}
%============================================================================================================================================================
\textbf{Databases and Protocols}: The Polathermal dataset \cite{hu2016polarimetric}, offers polarimetric LWIR imagery alongside color images for a total of 60 participants. We employed the five-fold protocols suggested in \cite{de2018heterogeneous} and presented the average Rank-1 identification rate across the five folds. The Tufts Face Database \cite{panetta2018comprehensive}, consists of face images captured through different modalities, aimed for the \hfr task. For assessing VIS-Thermal \hfr performance, we utilized the thermal images from this dataset, covering 113 identities. We follow the standard protocols in \cite{fu2021dvg} for reporting results. The SCFace dataset \cite{grgic2011scface}, comprises high-quality images suitable for face recognition during enrollment, whereas the probe samples are of lower quality, originating from surveillance cameras at different distances (heterogeneity in terms of image quality). This dataset contains four protocols, categorized according to the probe sample quality and distance: close, medium, combined, and far, with the "far" protocol representing the most challenging scenario.\\
\textbf{Performance Metrics}: We evaluate the models based on several performance metrics frequently used in recent literature, such as Area Under the Curve (AUC), Equal Error Rate (EER), Rank-1 identification rate, and Verification Rate at specific false acceptance rates (0.01\%, 0.1\%, 1\%, and 5\%).
\subsection{Experimental results}
%----------------------------------------------
The experiments conducted on various datasets and their results are elaborated in this section. For benchmarking, we contrast our results with those of the state-of-the-art, specifically compared with the methods presented in \cite{george2022prepended, george2023bridging}.\\
%%%----------------------------------------------
\textbf{Experiments with Polathermal dataset}: Experiments were performed in the `thermal-to-visible` recognition scenario in the Polathermal dataset, and the results are shown in Table \ref{tab:polathermal}. This table illustrates the mean Rank-1 identification rate across the five protocols of the Polathermal 'thermal to visible protocols' (following the reproducible protocols outlined in \cite{de2018heterogeneous}). The introduced \diu method attains a mean Rank-1 accuracy of 97.8\% with a standard deviation of 1.28\%, achieving the state-of-the-art performance.

\begin{table}[ht]
\caption{Pola Thermal - Average Rank-1 recognition rate}
\label{tab:polathermal}
\begin{center}
  \resizebox{0.6\columnwidth}{!}{
  \begin{tabular}{lr}
    \toprule
    \textbf{Method} & \textbf{Mean (Std. Dev.)} \\ \midrule
    
    DPM in \cite{hu2016polarimetric}   & 75.31 \% (-) \\ 
    CpNN in \cite{hu2016polarimetric}  & 78.72 \% (-) \\ 
    PLS in \cite{hu2016polarimetric}   & 53.05\% (-)  \\  \midrule

    LBPs + DoG in \cite{liao2009heterogeneous} & 36.8\% (3.5) \\ 
    ISV in \cite{de2016heterogeneous}       & 23.5\% (1.1) \\ 

    DSU(Best Result) \cite{de2018heterogeneous} & 76.3\% (2.1) \\

    DSU-Iresnet100 \cite{george2022prepended} & 88.2\% (5.8) \\
    % \hline
    % \rowcolor{Gray}
    PDT \cite{george2022prepended} & 97.1\% (1.3) \\ 
    
    CAIM \cite{george2023bridging} & 95.0\% (1.63) \\ \midrule

    \rowcolor{Gray}

    \textbf{DIU (Proposed)} & \textbf{97.8\%(1.28)}  \\
    \bottomrule 
  \end{tabular}
  }
\end{center}
% \vspace{-5mm}

\end{table}

%%%----------------------------------------------
\textbf{Experiments with Tufts face dataset}: Table \ref{tab:tufts} showcases the performance of the \diu method in comparison to other state-of-the-art methods in the VIS-Thermal protocol of the Tufts face dataset. This dataset poses significant challenges due to variations in pose and additional factors. Extreme yaw angles in the dataset lead to a performance dip for even the visible spectrum face recognition systems, paralleled by a decline in \hfr performance. Nevertheless, the \diu method achieves the highest verification rate (85.9\% at 1\% FAR) and a Rank-1 accuracy of 82.94\%. These results demonstrate the robustness of the proposed approach.
% \vspace{-5mm}

\begin{table}[h]
  \centering
  \caption{Experimental results on VIS-Thermal protocol of the Tufts Face dataset.}
  \label{tab:tufts}
  \resizebox{0.95\columnwidth}{!}{
  \begin{tabular}{lccc}
    \toprule
    \textbf{Method} & \textbf{Rank-1} & \textbf{VR@FAR=1$\%$} & \textbf{VR@FAR=0.1$\%$}  \\ \midrule
      LightCNN \cite{Wu2018ALC} & 29.4 & 23.0 & 5.3 \\
      DVG \cite{fu2019dual} & 56.1 & 44.3 & 17.1 \\
      DVG-Face \cite{fu2021dvg} & 75.7 & 68.5 & 36.5 \\ 
      DSU-Iresnet100 \cite{george2022prepended} & 49.7 & 49.8 & 28.3 \\   
      
      PDT \cite{george2022prepended} & 65.71 & 69.39 & 45.45 \\ 
      
      CAIM \cite{george2023bridging}  & 73.07 & 76.81 & 46.94 \\\midrule

      \rowcolor{Gray}

      \textbf{DIU (Proposed)} & \textbf{82.94} & \textbf{85.9} & \textbf{74.95}  \\

      \bottomrule
  
  \end{tabular}
  }
  % \vspace{-5mm}

\end{table}

\textbf{Experiments with SCFace dataset}: We performed experiments on the SCFace dataset in the visible images protocol. This dataset poses a challenge due to the quality difference between gallery (high-res mugshots) and probe (low-res surveillance camera) images. We report the results on the most challenging ``far" protocol with very low-quality probe images. From Table \ref{tab:scface}, it can be seen that the \diu approach achieves the highest Rank-1 accuracy at 94.55\%. This shows that the proposed approach can address heterogeneity in terms of image quality as well, meaning it can even be used to improve the performance of pretrained models. \\
% \vspace{-7mm}
\begin{table}[h]
  \caption{Performance of the proposed approach in the SCFace dataset. }
  \label{tab:scface}
  \centering
  \resizebox{0.98\columnwidth}{!}{%
  \begin{tabular}{lcrrrr}
  \toprule
  \textbf{Protocol}             & \textbf{Method} & \textbf{AUC}   & \textbf{EER}   & \textbf{Rank-1}    & \begin{tabular}[c]{@{}c@{}} \textbf{VR@}\\\textbf{FAR=0.1\%} \end{tabular} \\ \midrule

\multirow{3}{*}{Far}     
                                % \rowcolor{Gray}
                                &DSU-Iresnet100 \cite{george2022prepended} & 97.18 & 8.37 & 79.53 & 58.26  \\

                             & PDT \cite{george2022prepended}   &  98.31 & 6.98 &  84.19 & 60.00            \\ 
                             &  CAIM \cite{george2023bridging} &   98.81 &   5.09 &   86.05 &   61.86 \\  \cmidrule{2-6}

                             &\cellcolor{Gray}\textbf{DIU (Proposed)} & \cellcolor{Gray} \textbf{99.65} & \cellcolor{Gray}\textbf{2.73} & \cellcolor{Gray}\textbf{94.55} &\cellcolor{Gray} \textbf{82.73}\\

  \bottomrule
  \end{tabular}
  }
  % \vspace{-2mm}

  \end{table}
\textbf{Influence of the Number of DIU Layers}: To understand the impact of varying the number of adaptable DIU layers, we performed an experiment on the Polathermal dataset, with different number of DIU layers. The results are shown in Tab. \ref{tab:ablation_layer}. Initially, it can be seen that adapting a greater number of layers enhances performance. However, this improvement plateaus at a certain point, after which performance begins to decline. Among the 48 layers, the best results were achieved when unfreezing up to the 24th layer.
% \vspace{-3mm}

\begin{table}[!htb]
  \caption{Performance with different number of DIU blocks. 1-6 indicates the blocks from first to sixth layers are adapted. Experiment performed in Polathermal dataset. }
  \centering
  \resizebox{0.99\columnwidth}{!}{%
\begin{tabular}{llllll}
\toprule
\textbf{Layers} &             \textbf{AUC} &             \textbf{EER} &              \textbf{Rank-1} &          \textbf{VR (0.1\% FAR)} &            \textbf{VR (1\% FAR)} \\
\midrule
1 &  94.42$\pm$1.41 &  13.27$\pm$2.13 &  70.26$\pm$3.60 &  36.30$\pm$4.98 &  58.80$\pm$4.28 \\
1-6 &  98.64$\pm$0.56 &  6.00$\pm$1.30 &  88.29$\pm$1.78 &  60.57$\pm$3.26 &  79.64$\pm$5.29 \\
1-12 &  99.56$\pm$0.10 &  3.37$\pm$0.43 &  95.79$\pm$1.42 &  73.06$\pm$6.36 &  91.84$\pm$1.07 \\
1-18 &  99.77$\pm$0.14 &  2.22$\pm$0.94 &  \textbf{97.33$\pm$1.08} &  85.75$\pm$7.21 &  95.92$\pm$2.44 \\
\rowcolor{Gray}
1-24 &  \textbf{99.84$\pm$0.11} &  \textbf{2.06$\pm$0.87} &  96.88$\pm$2.11 &  88.47$\pm$2.56 &  \textbf{95.97$\pm$2.20} \\
1-30 &  99.75$\pm$0.18 &  2.12$\pm$1.19 &  96.74$\pm$2.43 &  \textbf{89.07$\pm$7.15} &  95.10$\pm$4.37 \\
1-36 &  99.68$\pm$0.23 &  2.48$\pm$1.11 &  96.04$\pm$1.95 &  77.39$\pm$9.28 &  93.98$\pm$3.92 \\
1-42 &  99.71$\pm$0.29 &  2.33$\pm$1.69 &  96.77$\pm$2.83 &  81.01$\pm$12.41 &  94.98$\pm$5.18 \\
1-48 &  99.64$\pm$0.17 &  2.80$\pm$0.69 &  96.01$\pm$2.47 &  70.12$\pm$14.50 &  89.30$\pm$6.27 \\
\bottomrule
\end{tabular}
  }
  \label{tab:ablation_layer}
\end{table}

\textbf{Impact of $\gamma$ }: The hyper-parameter $\gamma$ integrates the role of contrastive learning in aligning embeddings with the guidance of the teacher network during the distillation. To understand the influence of this parameter, we conducted experiments on the Polathermal dataset, examining how varying $\gamma$ values affect the resulting HFR performance. As evidenced by Tab. \ref{tab:ablation_gamma}, optimal performance is reached at $\gamma=0.75$. Owing to the adaptation of only a select number of lower layers, DIU still achieves reasonable performance at $\gamma=0$, which corresponds to supervision solely via contrastive loss. Incorporating the teacher's supervision further improves the results, achieving a peak AUC of 99.80\%.

\begin{table}[!htb]
  \caption{Performance with different values of $\gamma$. Experiment performed in Polathermal dataset. }
  \centering
  \resizebox{0.99\columnwidth}{!}{%
\begin{tabular}{llllll}
\toprule
\textbf{$\gamma$} &             \textbf{AUC} &             \textbf{EER} &              \textbf{Rank-1} &          \textbf{VR (0.1\% FAR)} &            \textbf{VR (1\% FAR)} \\
\midrule
0.00 &  99.75$\pm$0.10 &  2.07$\pm$0.57 &  97.13$\pm$0.66 &  66.97$\pm$22.62 &  94.86$\pm$2.23 \\
0.25 &  99.72$\pm$0.23 &  2.17$\pm$1.58 &  96.82$\pm$3.01 &  80.91$\pm$13.14 &  93.92$\pm$6.35 \\
0.50 &  99.77$\pm$0.16 &  2.25$\pm$1.27 &  96.27$\pm$2.37 &  85.49$\pm$7.64 &  94.15$\pm$4.60 \\
\rowcolor{Gray}
\textbf{0.75} &  \textbf{99.80$\pm$0.16} &  \textbf{2.02$\pm$1.06} &  \textbf{97.82$\pm$1.28} &  \textbf{87.50$\pm$6.85} &  \textbf{96.34$\pm$2.53} \\
1.00 &  92.45$\pm$1.41 &  16.88$\pm$2.13 &  69.23$\pm$2.71 &  33.10$\pm$1.51 &  50.22$\pm$3.70 \\
\bottomrule
\end{tabular}

  }
  \label{tab:ablation_gamma}
\end{table}

\textbf{Effect of FR architecture}: To compare the effectiveness of another architecture we considered the Iresnet50 model which is much smaller than the Iresnet100 model. As shown in Table \ref{tab:fr_arch}, the performance is better for the bigger model for the HFR task (as well as for FR), however the small model still achieved good HFR performance showing the approach can be extended to both small and large FR models.
\vspace{-2mm}
\begin{table}[h]
  \centering
  \caption{Comparison of FR architectures on Tufts Face dataset.}
  \label{tab:fr_arch}
  \resizebox{0.95\columnwidth}{!}{
  \begin{tabular}{lccc}
    \toprule
    \textbf{Method} & \textbf{Rank-1} & \textbf{VR@FAR=1$\%$} & \textbf{VR@FAR=0.1$\%$}  \\ \midrule

      DIU (Iresnet100) & \textbf{82.94} & \textbf{85.9} & \textbf{74.95}  \\

      DIU(Iresnet50) & 73.79 & 75.88 & 48.24 \\

      \bottomrule
  
  \end{tabular}
  }
  %  \vspace{-5mm}

\end{table}

\section{Conclusions}
\label{sec:conclusions}
%============================================================================================================================================================
In this work, we introduce a novel approach for learning domain-invariant layers called domain invariant units (DIU) for the challenging task of heterogeneous face recognition. Leveraging guidance from a pretrained face recognition network, our method successfully extracts domain-invariant features while optimizing a contrastive learning objective. Our extensive experiments have showcased state-of-the-art performance achieved by our proposed method across a range of challenging benchmark datasets, demonstrating its effectiveness and robustness.  The source codes and protocols will be made publicly available to facilitate the extension of our work.

\bibliographystyle{IEEEbib}

{\small \bibliography{refs}}

\end{document}